\PassOptionsToPackage{pdftex,dvipsnames}{xcolor}
\documentclass[letterpaper, 10 pt, conference]{ieeeconf}  

\IEEEoverridecommandlockouts                             
\usepackage{graphics}
\usepackage{epsfig} 
\usepackage{mathptmx} 
\usepackage{times} 
\usepackage{amsmath} 
\usepackage{amssymb}  
\usepackage{subcaption}
\usepackage{graphicx}
\usepackage{float}
\usepackage{bm}
\usepackage{caption}
\usepackage{mathtools}
\usepackage{stfloats}
\usepackage{mathrsfs}
\usepackage{multirow}
\usepackage{makecell}
\usepackage{vcell}
\usepackage{booktabs}
\usepackage{diagbox}
\usepackage{csquotes}
\usepackage{cite}
\usepackage{tabularx}
\usepackage{tikz}
\usepackage{lipsum}
\usepackage{schemata}
\usepackage{verbatim}
\usepackage{pifont}

\usepackage[hyphens]{url}
\usepackage[hidelinks]{hyperref}
\hypersetup{breaklinks=true}
\usepackage{cite}

\usepackage[ruled,vlined]{algorithm2e}
\SetKwInput{Input}{Input}
\SetKwInput{Output}{Output}
\SetKw{Function}{function }
\SetKw{Return}{return }
\SetKw{Returns}{\quad return }

\usepackage[noend]{algpseudocode}

\algrenewcommand\algorithmicforall{\textbf{for each}}
\algrenewcommand\algorithmicindent{.8em}

\algdef{SE}[DOWHILE]{Do}{doWhile}{\algorithmicdo}[1]{\algorithmicwhile\ #1}%
\algnewcommand\algorithmicforeach{\textbf{for each}}
\algdef{S}[FOR]{ForEach}[1]{\algorithmicforeach\ #1\ \algorithmicdo}

\usetikzlibrary{decorations.pathreplacing,calc}
\renewcommand{\Comment}[1]{\hfill\textcolor{ForestGreen}{\(\triangleright\) #1}}

\usepackage{hyperref}
\hypersetup{
colorlinks=true,
linkcolor=blue,
filecolor=magenta,
urlcolor=blue,
}
\usepackage[font=footnotesize]{caption}

\title{\LARGE \bf
Semantic Layering in Room Segmentation via LLMs
}
\author{Taehyeon Kim and Byung-Cheol Min
\thanks{The authors are with SMART Lab, Department of Computer and Information Technology, Purdue University, West Lafayette, IN 47907, USA \tt\small{\{kim4435, minb\}@purdue.edu}}%
}

\begin{document}
\maketitle
\thispagestyle{empty}
\pagestyle{empty}

\begin{abstract}

In this paper, we introduce Semantic Layering in Room Segmentation via LLMs (SeLRoS), an advanced method for semantic room segmentation by integrating Large Language Models (LLMs) with traditional 2D map-based segmentation. Unlike previous approaches that solely focus on the geometric segmentation of indoor environments, our work enriches segmented maps with semantic data, including object identification and spatial relationships, to enhance robotic navigation. By leveraging LLMs, we provide a novel framework that interprets and organizes complex information about each segmented area, thereby improving the accuracy and contextual relevance of room segmentation. Furthermore, SeLRoS overcomes the limitations of existing algorithms  by using a semantic evaluation method to accurately distinguish true room divisions from those erroneously generated by furniture and segmentation inaccuracies. The effectiveness of SeLRoS is verified through its application across 30 different 3D environments.
Source code and experiment videos for this work are available at: \url{https://sites.google.com/view/selros}.

\end{abstract}


\section{Introduction}
\label{sec:intro}
Navigating through home indoor environments with the aid of robotics has increasingly relied on vision-language cues for object-oriented navigation~\cite{anderson2018vision}\cite{zhu2020vision}\cite{murray2022following}\cite{obinata2023semantic}. Yet, the challenge of autonomously recognizing and targeting specific `contextual' places, such as kitchens, living rooms, and bathrooms, without direct human input remains largely unexplored. Traditional approaches to room segmentation have predominantly utilized 2D maps to classify spaces within indoor environments~\cite{bormann2016room}\cite{luperto2022robust}. However, these methods focus strictly on the segmentation aspect, neglecting the semantic information that is crucial for a more nuanced understanding and navigation of these spaces. This paper introduces \textbf{Se}mantic \textbf{L}ayering in \textbf{Ro}om \textbf{S}egmentation via LLMs (SeLRoS), an innovative approach that leverages Large Language Models (LLMs) to integrate semantic information into segmented 2D maps. This method significantly enhances the functionality of robotic navigation systems and the accuracy of traditional segmentation algorithms.

\begin{figure}[t]
\centering
\includegraphics[width=\columnwidth]{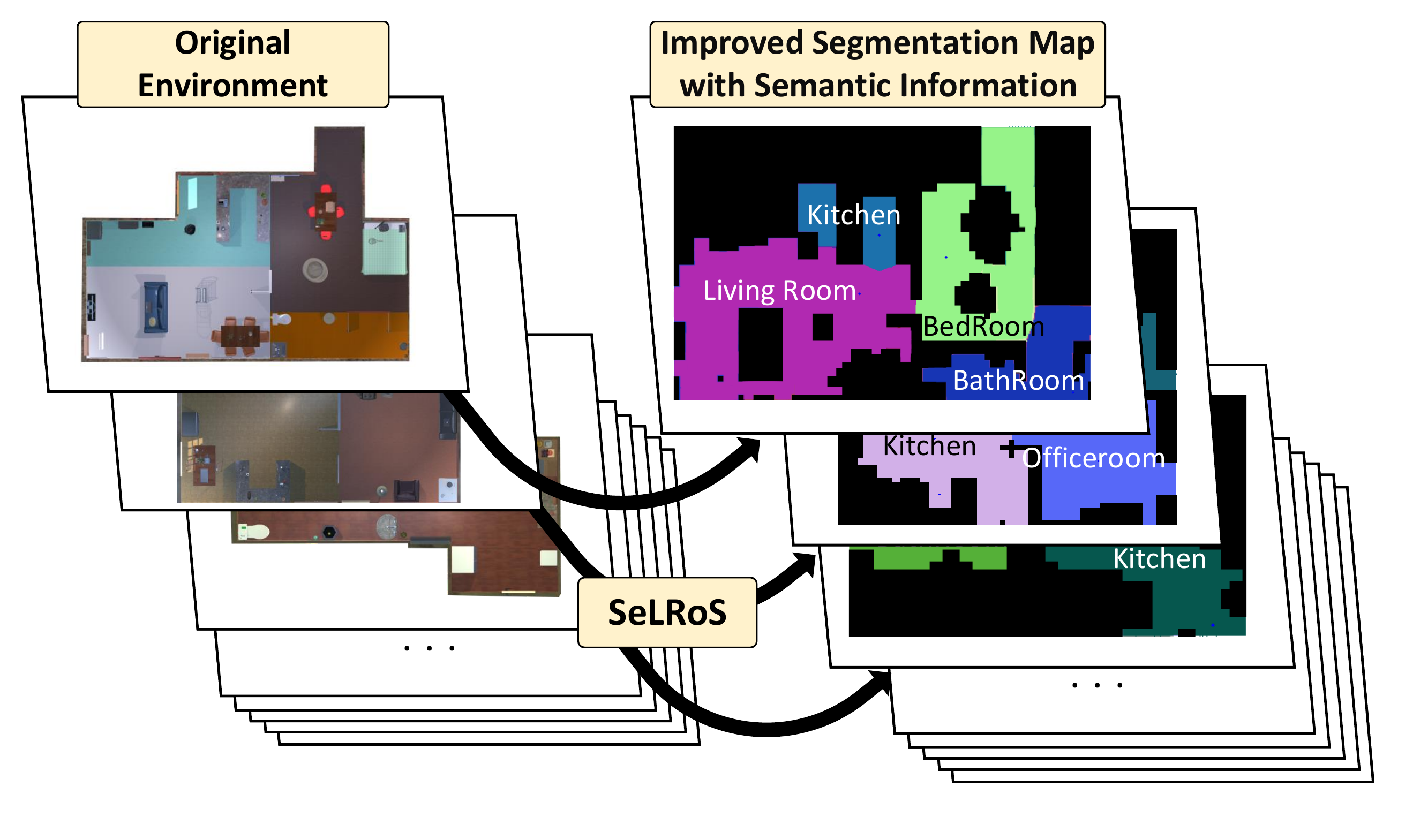}
\vspace{-20pt}
\caption{Semantic Layering in Room Segmentation via LLMs (SeLRoS) employs a room segmentation algorithm, an object detection algorithm, and Large Language Models (LLMs) to derive a 2D segmentation map from a 3D environment (left side of the figure), as well as to produce semantic information for each segmented room (right side of the figure).}
\vspace{-18pt}
\label{fig1}
\end{figure}

As illustrated in Fig.~\ref{fig1}, SeLRoS addresses the limitations of existing room segmentation algorithms by integrating semantic data into the segmentation process. This involves analyzing objects within each room, as well as considering spatial relationships and the physical characteristics of spaces, such as size and shape. By utilizing LLMs, we can organize this diverse array of information and make additional inferences, thus providing a richer, more contextually aware mapping of home indoor environments. The incorporation of semantic data not only augments the segmented maps with meaningful information but also serves to refine the segmentation algorithm itself.

Conventional room segmentation methods have shown proficiency in mapping unfurnished spaces~\cite{bormann2016room}\cite{luperto2022robust}\cite{8961806}. However, they falter when confronted with complex environments where furniture and other objects can lead to the erroneous division of rooms. By applying semantic insights, our study aims to discern whether segmented spaces are distinct rooms or erroneously divided sections of the same room due to the limitations of current algorithms. This semantic evaluation allows for a more accurate representation of indoor spaces, distinguishing between, for instance, multiple bedrooms or a single living room fragmented by the presence of furniture.

In summary, this paper introduces SeLRoS, a novel approach to room segmentation by leveraging the analytical prowess of LLMs to assign semantic information to 2D maps. This method not only enhances the accuracy of room segmentation in furnished environments but also enriches the semantic understanding of indoor spaces, paving the way for more sophisticated and contextually aware navigation systems.
The contributions of this paper are as follows:
\begin{itemize}

 \item We propose an innovative architecture using LLMs to integrate semantic information into existing room segmentation results including algorithm for interpreting room segmentation outcomes and a prompt engineering technique.

 \item We enhance the accuracy of room segmentation by utilizing semantic data to rectify segmentation errors caused by furniture in home indoor environments.

 \item We conducted extensive experiments to validate SeLRoS and have made the source code and related-map files available to the community for further research and development.
\end{itemize}


\begin{figure*}[t]
\includegraphics[width=\linewidth]{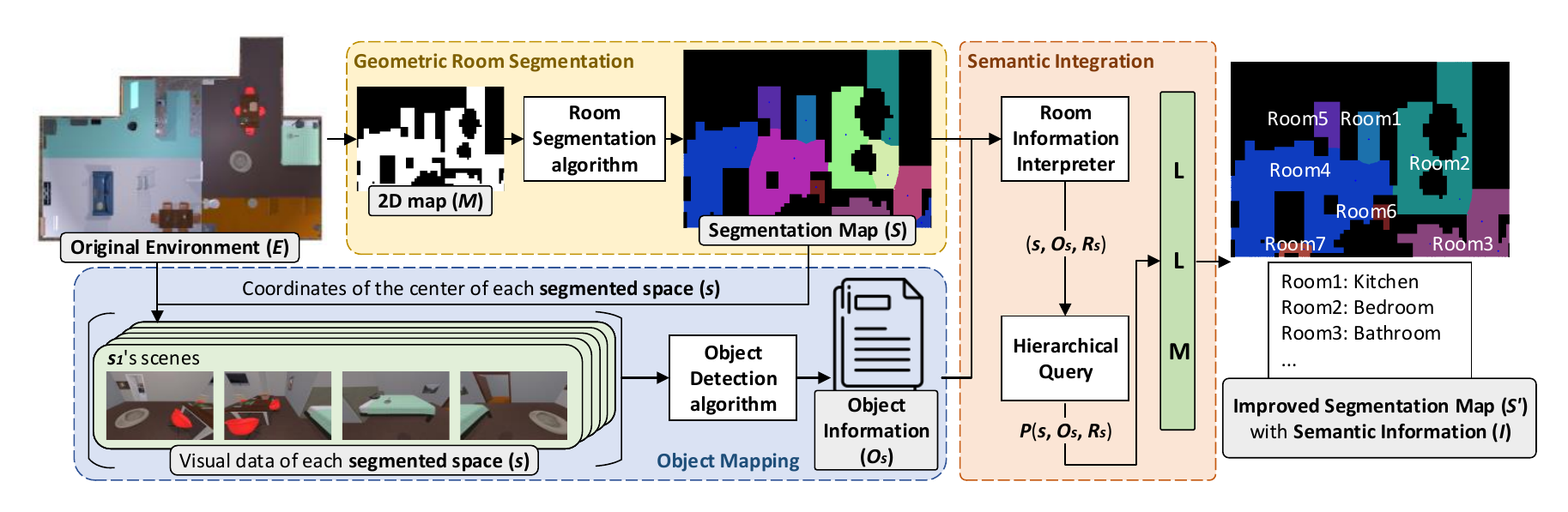}
\vspace{-15pt}
\caption{Overview of SeLRoS’s structure: SeLRoS begins with \textcolor{Goldenrod}{Geometric Room Segmentation}, where a 2D map (\(M\)) from the Original Environment (\(E\)) is transformed into a Segmentation Map (\(S\)). Following this, the \textcolor{Cerulean}{Object Mapping} process extracts Object Information (\(O_s\)) by analyzing scenes from the Original Environment’s center coordinates of each segmented space (\(s\)), employing an Object Detection algorithm. In the \textcolor{Orange}{Semantic Integration} process, harmonizing \(s\), \(O_s\) and the data of spatial relations (\(R_s\)) through the Room Information Interpreter and generating prompts \(P(s, O_s, R_s)\) via Hierarchical Query. The final outputs are Improved Segmentation Map (\(S'\)) with Semantic Information (\(I\)).}
\vspace{-10pt}
\label{fig2}
\end{figure*}

\section{Related Works}
\label{sec:rel_work}

\noindent\textbf{Traditional Room Segmentation.} Research in traditional room segmentation primarily focuses on utilizing 2D maps to delineate individual rooms within an indoor environment~\cite{bormann2016room}\cite{8461128}. Recent advancements in this field have achieved considerable accuracy in segmenting spaces based on geometric properties~\cite{luperto2022robust}. However, these methods often face limitations when applied to environments with furniture, where the presence of objects can significantly impact segmentation accuracy. To address these challenges, some studies have explored the use of additional sensors, such as 3D LiDARs, to enhance segmentation precision~\cite{8961806}\cite{7814251}\cite{8019484}. However, as suggested in this paper, the integration of semantic information into the segmentation process to improve the utility and accuracy of segmented maps in furnished environments has yet to be fully explored.

\noindent\textbf{Indoor Scene Classification.} Indoor scene classification research aims to identify and classify various scenes within indoor environments based on their characteristics and features~\cite{afif2020deep}\cite{basu2020indoor}\cite{9341580}\cite{zhou2021borm}\cite{miao2021object}. This approach differs from room segmentation in that it focuses on the scene's content rather than its geometric boundaries. While these studies provide valuable insights into understanding indoor environments\cite{labinghisa2022indoor}, their application to traditional robot navigation, particularly in enhancing segmented maps, is limited. Moreover, although scene classification can infer the function of a specific area, it does not address the challenge of defining the spatial extent of these inferred places, leaving a gap in the application of scene classification data to practical navigation tasks.

\noindent\textbf{Robotics-related Research Using LLMs.} The use of LLMs in robotics-related research represents a burgeoning area of interest, marked by various innovative attempts to integrate LLMs capabilities into robotic systems\cite{wu2023tidybot}\cite{zhang2023large}\cite{ding2023task}. While the potential of LLMs in enhancing robot autonomy and decision-making is vast, their application, especially in writing comprehensive robot control codes, poses stability and reliability concerns\cite{vemprala2023chatgpt}\cite{singh2023progprompt}\cite{Wang2023Prompt}\cite{cao2023ground}. Our study distinguishes itself within this context by leveraging the strengths of LLMs not to replace, but to augment existing methods\cite{shah2023navigation}\cite{yu2023scaling}\cite{yu2023l3mvn}. By combining LLMs with widely used segmentation techniques, our approach aims to enhance the semantic understanding of segmented spaces, thereby offering a novel contribution to the field of robotic navigation and indoor mapping.


\section{Problem Formulation}
\label{sec:ProbFomul}

Let \(S\) represent the set of all segmented spaces based on original indoor environment \(E\), where each space \(s \in S\) is defined by its geometric properties derived from a 2D map. The traditional room segmentation problem can be defined as a function \(f : M \rightarrow S\), where \(M\) represents the 2D map of the environment, and \(S\) is the set of segmented spaces. The semantic gap is defined by the lack of functional and contextual information in \(S\), which is necessary for distinguishing between different types of rooms beyond their geometric properties.

We aim to obtain semantic information \(I\) and utilize it to enhance Set \(S\), resulting in the creation of \(S'\). This semantic enrichment boosts the accuracy of the resulting set and expands the usability of the map.
Let \(O_s\) represent the set of objects observed in the segmented environment, and \(R_s\) represent the set of relations among segmented rooms \(s\), including adjacency and spatial characteristics. Given a segmented space \(s\) and its corresponding objects and relations \(O_s\) and \(R_s\), we formulate a prompt \(P(s, O_s, R_s)\) that encapsulates this information for the LLM. The LLM's response to \(P\) becomes semantic information \(I\). This \(I\) is used to improve \(s\) by providing additional context, transforming \(s\) into \(s'\).
Therefore, the objective of our study is to define a function \(g : S \times I \rightarrow S'\) that maps each segmented space \(s\), enhanced with semantic information \(I\) derived from \(P(s, O_s, R_s)\), to \(s'\).

\section{Methodology}
\label{sec:methodology}
SeLRoS consists of three parts: \textit{geometric room segmentation}, \textit{object mapping}, and \textit{semantic integration}, as depicted in Fig.~\ref{fig2}.

\subsection{Geometric Room Segmentation}
This process involves generating a segmentation map (\(S\)) from a 2D map (\(M\)), which itself is derived from the original environment (\(E\)). The purpose of this phase is to accurately delineate the spatial boundaries within an environment, setting the stage for further semantic enhancement.

For the implementation of SeLRoS, the Voronoi Random Field (VRF) algorithm is chosen for room segmentation. This decision is informed by a comparative analysis of existing renowned room segmentation algorithms~\cite{bormann2016room}\cite{luperto2022robust}. While morphological and distance algorithms are recognized for producing clear and well-defined segmentation maps, they tend to prioritize geometric accuracy over the nuanced understanding of space. Conversely, VRF adopts a variability and stochastic approach~\cite{vrf}. This method is adept at capturing more intricate segmentation patterns, offering a richer, albeit more fragmented, portrayal of room divisions. The challenge of over-segmentation associated with VRF, typically seen as a drawback, is adeptly mitigated in SeLRoS through the subsequent integration of semantic information. SeLRoS effectively consolidates excessively divided segmented spaces using semantic information.

An important aspect of SeLRoS's design is its modular architecture. This modularity ensures flexibility in the choice of \textit{Room Segmentation algorithms}. Should the need arise to adapt SeLRoS to different environments or to leverage advancements in segmentation techniques, the \textit{Room Segmentation algorithm} can be seamlessly updated or replaced.

\subsection{Object Mapping}
In the object mapping phase of the SeLRoS, we first determine the geometric center of each segmented room (\(s\)). This is achieved by calculating the mean position of all constituent points within a segment, specifically by averaging the $x$ and $y$ coordinates separately to ascertain the room's centroid.

Subsequently, we acquire visual data from the original environment (\(E\)) at these center positions. Utilizing an \textit{Object Detection algorithm}, we then systematically extract a list of objects present within each scene. For the purpose of object detection within SeLRoS, we employ Detic\cite{zhou2022detecting}, a robust algorithm known for its accuracy and efficiency in identifying and classifying objects within complex environments.

This procedure is executed for every segmented room, resulting in the creation of object information (\(O_s\)) for the segmentation map. This information specifies which objects are located in each segmented room, effectively mapping object presence across the entire segmented area.

\subsection{Semantic Integration}
The semantic integration process comprises two distinct parts. Initially, the \textit{Room Information Interpretation} phase processes the segmentation map (\(S\)), extracting key details for each segmented room, including area, shape, and adjacency relationships with neighboring rooms. This phase also integrates this spatial data with the object information~(\(O_s\)). Following this, the \textit{Hierarchical Query} stage implements prompt engineering techniques to efficiently relay this combined information to LLMs, ensuring reliable and high-quality outcomes.

\begin{figure}[t]
\includegraphics[width=\linewidth]{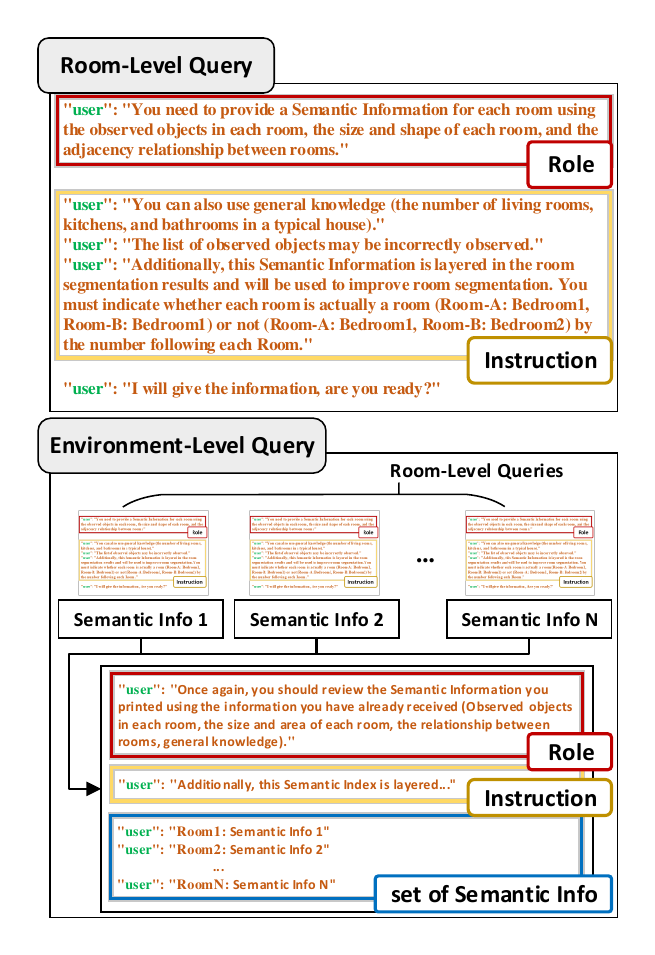}
\vspace{-20pt}
\caption{\textit{Hierarchical Query} is hierarchically composed of \textit{Room-Level Query} and \textit{Environment-Level Query}. The \textcolor{red}{red} box represents the role component, the \textcolor{Goldenrod}{yellow} box represents the instruction, and \textcolor{blue}{blue} box signifies the set of Semantic Information.}
\vspace{-18pt}
\label{fig3}
\end{figure}

\subsubsection{Room Information Interpretation}
The \textit{Room Information Interpretation} phase stands as a crucial element of our methodology, designed to analyze the segmentation map~(\(S\)) through the \textit{Room Information Interpreter}. This process entails a comprehensive interpretation of each segmented room's characteristics, including its area, shape, and the adjacency relationships it shares with surrounding rooms.
Initially, the interpreter identifies each room within the segmentation map~(\(S\)) by utilizing unique RGB values as markers. It then calculates the area of each room by tallying the pixels corresponding to each RGB value, offering an accurate representation of room sizes. To assess the shape of each room, our approach involves estimating the contours of rectangular shapes that approximate each room's boundaries, from which we derive the dimensions—length and width—providing a geometric approximation of the room's shape. Furthermore, the adjacency relationships between rooms are determined by slightly dilating each room's mask to create an expanded boundary. When these dilated masks overlap with those of neighboring rooms, it indicates adjacency, thereby enabling us to map the complex network of connections between rooms within the segmented environment.

\subsubsection{Hierarchical Query}
\textit{Hierarchical Query}, depicted in Fig.~\ref{fig3}, is our novel approach to prompt engineering. This technique is inspired by and extends upon concepts from prior research, drawing upon the structural query organization found in our previous work \cite{kim2023dynacon} and the analytical decomposition from \cite{wei2022chain}. However, \textit{Hierarchical Query} advances these ideas by partitioning a comprehensive query—aimed at requesting semantic information~(\(I\)) for each segmented room~(\(s\))—into a series of targeted sub-queries. These sub-queries, comprising both \textit{Role} and \textit{Instruction} elements, are individually presented to LLMs. The \textit{Role} is crafted to designate a specific function to the LLMs, guiding it on which domain of knowledge to access by setting a particular context or perspective. Concurrently, the \textit{Instruction} serves to define the bounds of the LLM's action. Considering the LLM's foundation on vast datasets for information, this \textit{Instruction} acts as a constraint, either by prescribing a desired format for structuring the answer or by narrowing the dataset's scope through a description of the current context.

The core distinction of \textit{Hierarchical Query} lies in its integration of responses. Initially, LLMs address the segmented \textit{Room-Level Query}, engaging with discrete elements of the environment. Subsequently, the responses to these sub-queries are aggregated, and the \textit{Environment-Level Query}, or in other words, the prompt~\(P(s, O_s, R_s)\) that is ultimately delivered to the LLMs, is generated. This process permits LLMs to reconsider their previous responses within the complete context information and enhances the accuracy and relevance of the answers.

A pivotal aspect of \textit{Hierarchical Query} is its emphasis on the format and structure of the responses. By incorporating instruction within the \textit{Environment-Level Query}, we guide LLMs to not only revisit their answers but also to align their responses with a predefined answer format. This prompt engineering technique ensures that the obtained answers are not only contextually enriched but also consistent and structured, significantly boosting the reliability and stability of the semantic information~(\(I\)).

\begin{algorithm}[t!]
\caption{Process of SeLRoS}\label{alg:selros}
\Input{Original Environment $E$} 
\Output{Improved Segmentation Map $S'$, Semantic Information $I$}

\SetKwBlock{DoParallel}{\Function{\upshape \Call{GeometricRoomSegmentation}{$E$}}}{}
    \DoParallel{
        $M \gets \Call{Get\_2D\_map}{E}$
        
        $S \gets \Call{VRF\_Segmetation}{M}$
    
        \For{\textbf{\upshape each} room \text{\upshape in} $S$}{
            $C[room] \gets \Call{Get\_center}{room}$
        }
    }
\Returns $S,C$

\SetKwBlock{DoParallel}{\Function{\upshape \Call{ObjectMapping}{$S, C$}}}{}
    \DoParallel{
        \For{\textbf{\upshape each} room \text{\upshape in} $S$}{
            $scene[room] \gets \Call{GetVisualData}{C[room]}$

            $O[room] \gets \Call{ObjectDetection}{scene[room]}$
        }
    }
\Returns $O$

\SetKwBlock{DoParallel}{\Function{\upshape \Call{SemanticIntegration}{$S, O$}}}{}
    \DoParallel{
        \For(\Comment{Room Info Interpretation}){\textbf{\upshape each} room \text{\upshape in} $S$}{
            
            $area \gets \Call{CalculateArea}{room}$

            $shape \gets \Call{CalculateShape}{room}$

            $adjacency \gets \Call{DetectAdjacency}{room}$

            $R[room] \gets \{area, shape, adjacency\}$
        }

        \For(\Comment{Room-Level Query}){\textbf{\upshape each} room \text{\upshape in} $S$}{
            
            $p[room] \gets \Call{MakePrompt}{O[room], R[room]}$

            $i[room] \gets \Call{LLM}{p[room]}$
        }

    $P \gets \Call{MakePrompt}{i, O, R}$\Comment{Env-Level Query}

    $I \gets \Call{LLM}{P}$

    $S' \gets \Call{IntegrationMap}{S, I}$
        
    }
\Returns $S', I$

$S, C \gets \Call{GeometricRoomSegmentation}{E}$

$O \gets \Call{ObjectMapping}{S, C}$

$S', I \gets \Call{SemanticIntegration}{S, O}$

\Return $S', I$
\end{algorithm}

The pseudo-code, detailed in Algorithm~\ref{alg:selros}, provides a technical overview of the entire SeLRoS process. This methodology significantly enhances the segmentation map by incorporating semantic information, thereby allowing for a more accurate and context-aware delineation of spaces within the environment.


\section{Experiments}
\label{sec:experiments}

\subsection{Experimental Setup}
To demonstrate SeLRoS's applicability across a spectrum of structures, experiments were conducted in various home indoor environments, utilizing 30 diverse 3D models generated through ProcTHOR\cite{procthor} within the AI2-THOR\cite{ai2thor} framework.
For each environment, a 2D map was crafted using AI2-THOR's reachable position extraction function, from which a segmentation map was subsequently derived. Following the procedures outlined in Section \ref{sec:methodology}, visual data were captured from the center coordinates of each segmented room, serving as the basis for acquiring object information. This collected data was processed through the semantic integration phase, resulting in an improved segmentation map enriched with semantic information. 

\subsection{Evaluation Criteria}
To substantiate our contributions, evaluation is bifurcated into qualitative and quantitative analyses. The qualitative analysis involves a heuristic comparison of our result against the original 3D environment, providing an intuitive assessment of the improvements offered by SeLRoS. 

The quantitative analysis is further split into two parts. The initial part evaluates the enhancements in room segmentation accuracy brought about by SeLRoS, utilizing well-known room segmentation algorithms as baselines for comparison. For this evaluation, we employ Intersection over Union~(IoU) and our newly introduced evaluation criterion, Match Scaled Intersection over Union (MSIoU), to quantitatively assess these improvements. The IoU metric, traditionally used to gauge segmentation accuracy, measures the overlap ratio between the predicted segmentation and the ground truth versus their collective area. MSIoU, however, enhances this assessment by implementing a scaling mechanism that adjusts based on the match quality rank of each segmented room's correspondence. The MSIoU formula is articulated as follows:
\begin{equation}
\text{MSIoU} = \text{$\frac{1}{N} \sum_{i=1}^{N} \left( \sum_{j=1}^{M_i} \text{IoU}(A_i, S_{ij}) \times \Delta\alpha \right)$}
\end{equation}
where \(N\) represents the total number of actual rooms, while \(M_i\) indicates the number of segmented rooms associated with the actual room \(A_i\). The \(\text{IoU}(A_i, S_{ij})\) refers to the IoU for the \(j\)th best-matched segmented room \(S_{ij}\) relative to the actual room \(A_i\). The decrement factor, \(\Delta\alpha\), is applied to each successive match's IoU value to diminish its weight, starting from the best match. In this context, \(\Delta\alpha\) decreases from 1.0 by 0.1 for each subsequent match, with a minimum value of 0.1, ensuring that even the lowest-ranked matches contribute to the overall MSIoU but with diminishing influence.

This metric not only allows us to assess the geometric accuracy of the segmentation but also introduces a graded evaluation that reflects the segmentation performance more comprehensively, taking into account both accuracy and the algorithm's tendency towards over-segmentation. Through this dual approach, utilizing both IoU and MSIoU, our analysis aims to provide a more detailed and nuanced understanding of SeLRoS's contributions to the field of room segmentation.

\begin{figure*}[t]
\includegraphics[width=\linewidth]{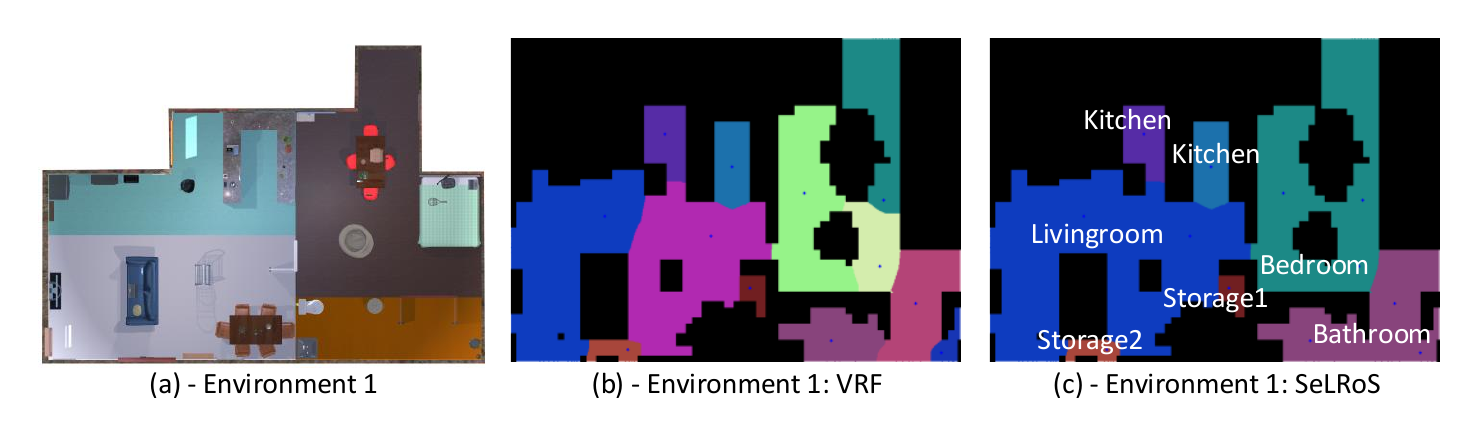}
\vspace{-20pt}
\caption{Results for Environment 1 - (a) depicts the original 3D environment, (b) shows the segmentation map created using the Voronoi Random Field (VRF) algorithm, and (c) presents the improved segmentation map, the final result achieved through SeLRoS, with semantic information added for readability.}
\vspace{-10pt}
\label{fig4}
\end{figure*}

The subsequent part, combining of baseline and ablation studies, focuses on evaluating the accuracy of the semantic information generated by SeLRoS. This is achieved by benchmarking against existing indoor scene classifier and examining the performance impact of omitting specific methodologies from our proposed approach. Through this comprehensive review, we aim to underscore the validity and innovative capacity of SeLRoS.


\section{Results and Analysis}
\label{sec:results}

\subsection{Qualitative Analysis}
In this section, we delve into the visual and interpretative evaluation of the improved segmentation map with semantic information outputted by our proposed SeLRoS system. This analysis focuses on observing the enhancements achieved through SeLRoS in comparison to the initial 3D environments and the segmentation maps generated by the VRF algorithm. Our experiments conducted across 30 3D environments, from which two environments will be selected for detailed review and analysis in the subsequent sections. All experiment results for the 30 environments can be found at our project website at: \url{https://sites.google.com/view/selros}. This qualitative assessment serves to highlight the tangible benefits of incorporating semantic information into the segmentation map, showcasing the practical application and effectiveness of SeLRoS in refining and enriching segmentation maps for enhanced spatial understanding and recognition.

\begin{figure*}[t]
\includegraphics[width=\linewidth]{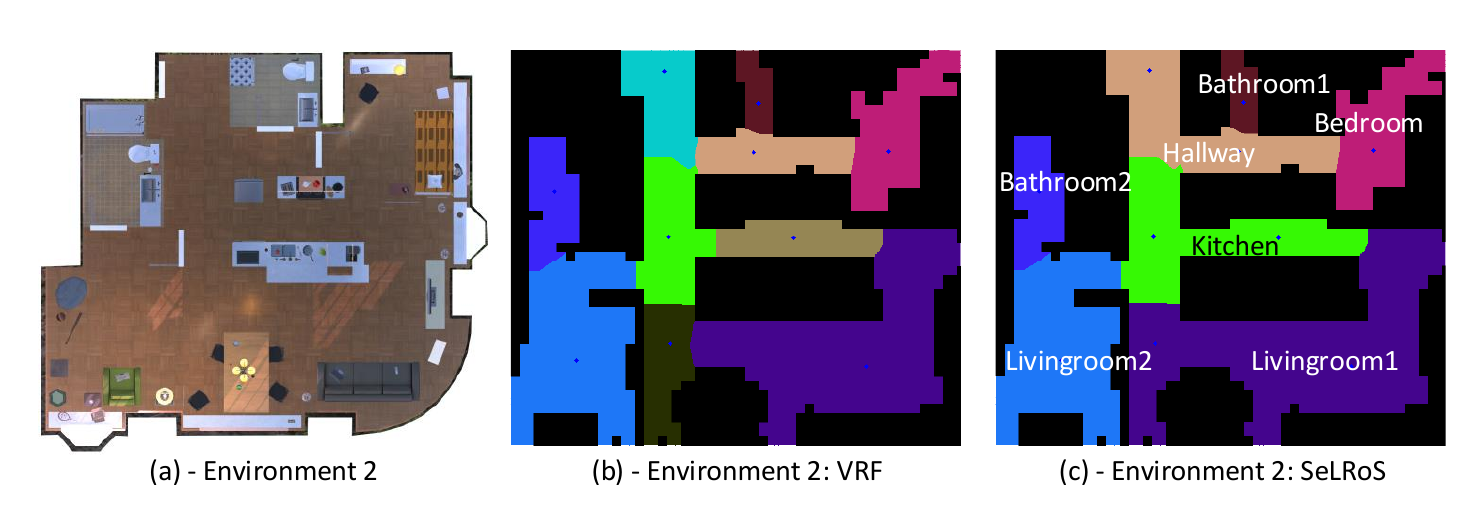}
\vspace{-20pt}
\caption{Results for Environment 2 - (a) depicts the original 3D environment, (b) shows the segmentation map created using the VRF algorithm, and (c) presents the improved segmentation map, the final result achieved through SeLRoS, with semantic information added for readability.}
\vspace{-10pt}
\label{fig5}
\end{figure*}

\subsubsection{Environment 1}
In the qualitative analysis of Environment 1, the outcomes are showcased in Fig.~\ref{fig4}, highlighting the effectiveness of SeLRoS in enhancing room segmentation through semantic integration. Initially, the application of SeLRoS is evident in the consolidation of segmented spaces that were previously divided without necessity. A notable example is the integration of areas marked in blue and purple on the center-right of the map, now collectively identified as the Livingroom, thanks to specified semantic information. Furthermore, a region in the upper right corner, initially segmented into three separate areas, has been unified under the singular semantic information of Bedroom. Similarly, what were once three distinct segmented rooms within the bathroom area at the bottom right have been merged into a single area. This consolidation underscores the capability of SeLRoS to utilize the relational data between observed objects and their corresponding spaces, effectively addressing the issue of rooms fragmented by the presence of furniture when relying solely on 2D map-based segmentation.

With regard to semantic information, the application of SeLRoS has enabled precise room identification: the Livingroom is designated where the TV and sofa reside, the Bedroom is identified by the presence of a bed, and the Bathroom is recognized by the toilet's location. Additionally, the Kitchen area, discernible near the sink in the 3D environment, illustrates a nuanced aspect of semantic integration. However, the case of kithchen, the initial segmentation inaccurately divided this area into two separate sections with intervening Livingroom space in the VRF-derived segmentation map, preventing their integration. Also, some misclassifications were observed in the Storage area. Despite their open-space nature, these areas were incorrectly labeled as storage by LLMs, a judgment likely influenced by the small size of the segmented spaces and the diversity of observed objects within them.

\subsubsection{Environment 2}
The analysis of Environment 2, as illustrated in Fig.~\ref{fig5}, also reveals significant advancements in the segmentation of indoor environments through the SeLRoS. In this instance, it was observed that segmented rooms, previously divided due to the presence of furniture such as the Hallway, Kitchen, and Livingroom1, were effectively integrated through semantic information.

Regarding the precision of the semantic information, most areas were accurately categorized, except for a space labeled as Livingroom2 located at the bottom left. This particular room, more appropriately characterized as an Officeroom based on its function and contents, was inaccurately identified as another living room. This likely stems from this room's size being significantly larger than what is typically observed for bedrooms.

Through the experiment, the utility of SeLRoS in enhancing room segmentation by integrating semantic information is further validated, demonstrating its potential in overcoming challenges posed by traditional segmentation approaches. However, the mislabeling of the Officeroom as Livingroom2 underscores the need for refined criteria within LLMs that consider beyond mere spatial dimensions to include a broader range of contextual information for accurate semantic labeling.

\subsection{Quantitative Analysis}

\subsubsection{Improvement in Segmentation Accuracy}
In our first part of quantitative analysis, the segmentation accuracy of SeLRoS is compared against four conventional segmentation algorithms: Morphological Segmentation\cite{mor}, Distance Segmentation\cite{dist}, Voronoi Segmentation\cite{voronoi}, and VRF Segmentation\cite{vrf}.

Morphological Segmentation, Distance Segmentation, Voronoi Segmentation, and VRF Segmentation each bring distinct approaches to room segmentation, exhibiting varying degrees of effectiveness across different environmental setups. Morphological Segmentation shines in settings with clear structural delineations for room boundaries but faces challenges in more intricate scenes where furniture and other items blur these edges. Distance Segmentation leverages proximity measures to differentiate spaces, yet its accuracy diminish in environments densely populated with furniture similarly to Morphological Segmentation. Meanwhile, Voronoi-based methods, including both Voronoi Segmentation and VRF Segmentation, apply a mathematical strategy for dividing space that heavily relies on the strategic placement of seed points. While this approach is adept at sketching out a basic spatial layout, it is susceptible to over-segmenting areas, inadvertently increasing the perceived number of rooms. 

\begin{figure}[t]
\includegraphics[width=\linewidth]{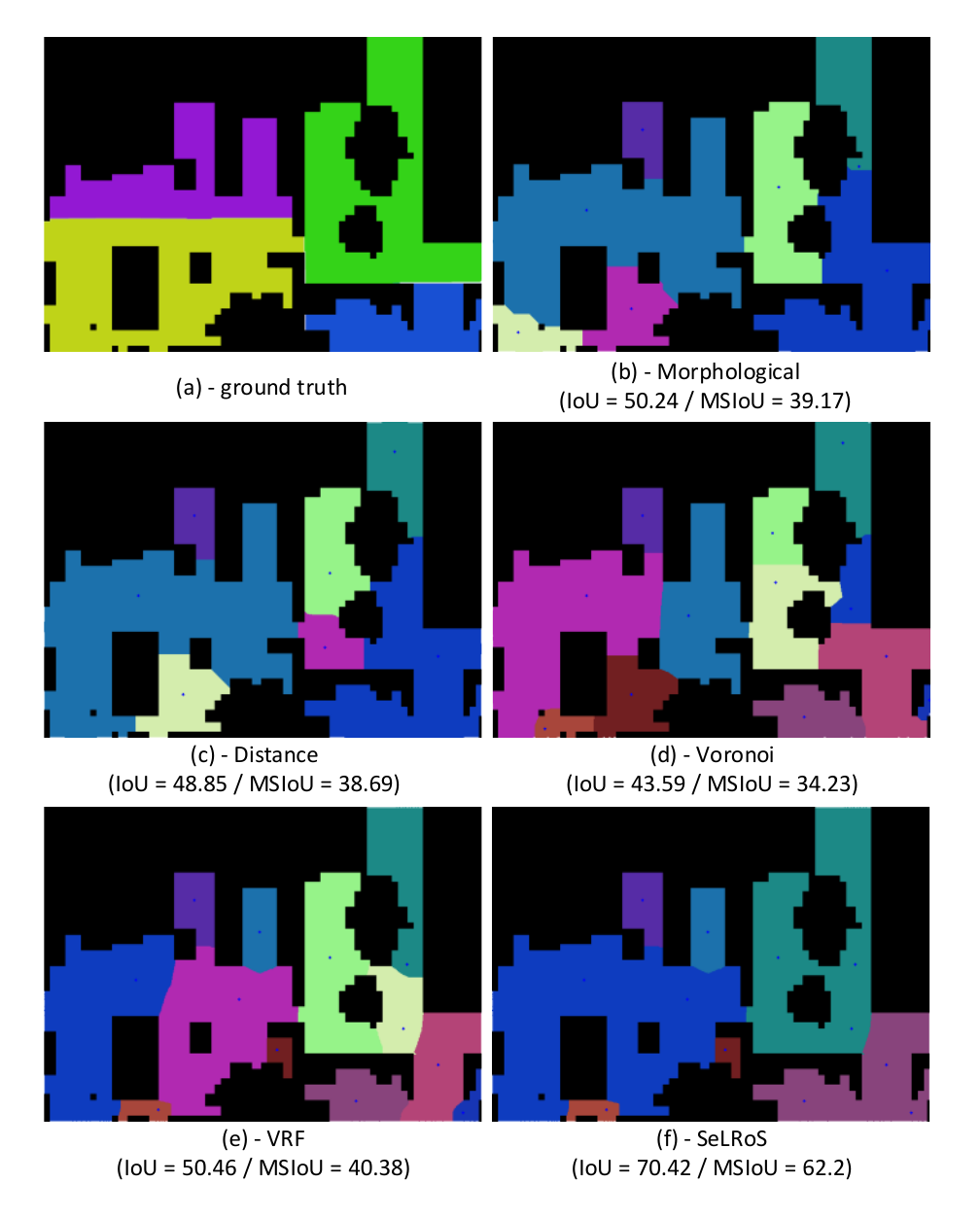}
\vspace{-15pt}
\caption{Comparison results for the Environment 1, showing the performance of four existing segmentation algorithms (Morphological, Distance, Voronoi, VRF) from (b) to (e), and SeLRoS (f). This is done by comparing them using IoU and MSIoU metrics with the ground truth (a).}
\vspace{-10pt}
\label{fig6}
\end{figure}

Fig.~\ref{fig6}, which presents the segmentation results alongside the ground truth for Environment 1 upon applying each segmentation algorithm, elucidates the distinctive behaviors of these algorithms. For the Morphological and Distance Segmentation algorithms, the segmentation results in expansive areas like the Livingroom align closely with those of SeLRoS, showing no superfluous segmentation. However, inaccuracies arise in the Bedroom, especially around a table, leading to erroneous segmentation. Conversely, the two Voronoi-based segmentation approaches demonstrate improved segmentation performance in the Bathroom area at the right-bottom, surpassing the other algorithms in this aspect. Yet, they tend to generate an excessive number of segmented rooms.

\begin{table}[h]
\caption{Experiment results for room segmentation accuracy}
\vspace{-10pt}
\begin{center}
\begin{tabular}{|c|c|c|c|c|c|}
\hline
\cline{1-5} 
\hline
\text{Test} & \text{Morphological} & \text{Distance}& \text{Voronoi}& \text{VRF} & \textbf{SeLRoS} \\
\hline
IoU & 63.27 & 59.45 & 60.51 & 61.89 & \textbf{69.98} \\
MSIoU & 52.36 & 49.41 & 50.7 & 51.56 & \textbf{57.73} \\
\hline
\end{tabular}
\label{tab1}
\end{center}
\vspace{-12pt}
\end{table}

Our experiments for comparison is grounded on the evaluation metrics of IoU and MSIoU, with the detailed numerical results presented in Table~\ref{tab1}, covering experiments across 30 different environments. The analysis confirms SeLRoS's superior performance in both IoU and MSIoU metrics, indicative of its enhanced precision in identifying room boundaries. Specifically, SeLRoS's capability to integrate semantic information effectively mitigates the over-segmentation tendency seen in Voronoi-based methods. Through this comparative analysis, it becomes evident that the integration of semantic information is pivotal in transcending the limitations inherent to purely geometric approaches to room segmentation.

\subsubsection{Evaluating Semantic Information}

In the second part of quantitative analysis, we conducted a comparative evaluation focusing on the semantic information derived from scene data observed at the center of each segmented room. The findings from this analysis are detailed in Table~\ref{tab2}. For benchmarking purposes, we utilized an Indoor Scene Classifier, which processes images to categorize indoor environments, as the baseline for our study. For this, we leverage YOLOv8 Image Classifier \cite{UltralyticsYOLOv8}, trained on the MIT Indoor Scene Recognition Image Dataset (comprising 15k images) \cite{mitindoor}.

\begin{table}[h]
\caption{Experiment results for semantic information accuracy}
\vspace{-10pt}
\begin{center}
\begin{tabular}{|c|c|c|c|c|}
\hline
\cline{1-5} 
\hline
\text{Test} & \text{\begin{scriptsize} Scene Classifier\end{scriptsize}} & \text{\begin{scriptsize}Obj Info\end{scriptsize}}& \text{\begin{scriptsize}Obj \&\ Room Info\end{scriptsize}}& \textbf{\begin{scriptsize}SeLRoS\end{scriptsize}} \\
\hline
Acc (\%) & 62.67 & 67.5 & 72.57 & \textbf{82.26} \\
\hline
\end{tabular}
\label{tab2}
\end{center}
\vspace{-5pt}
\end{table}

Additionally, we engaged in an ablation study by examining variations of our proposed method, where specific components were systematically omitted to assess their impact on performance. The ablation study differentiates between two key variations: \textit{Obj Info}, which represents the semantic information (\(I\)) generated from only the object information~(\(O_s\)) observed at the center of each segmented room, and \textit{Obj \& Room Info}, which represents the semantic information (\(I\)) created by object information (\(O_s\)) and contextual information (\(R_s\)) interpreted by the Room Information Interpreter. Notably, these variations exclude the application of SeLRoS's hierarchical query prompt engineering technique.

Our results demonstrate that the semantic information generated by SeLRoS outperforms all other methods under comparison in terms of accuracy. A significant observation is the underperformance of the Indoor Scene Classifier based on YOLOv8 compared to its typical application. This discrepancy arises from our unique methodology for capturing scene data, which diverges from conventional indoor scene classification that relies on scenes rich in distinctive objects. In our approach, which gathers images from four directions centered within each segmented room, may capture scenes dominated by non-descriptive elements like walls and doors, especially in smaller areas. This scenario can affect the classifier's performance, implicating its limited applicability to the SeLRoS framework, which employs a more comprehensive strategy for semantic information extraction by considering the entire environment's context.


\section{Conclusion}
\label{sec:conclusion}
In this paper, we introduced SeLRoS, a novel framework designed to enhance room segmentation with semantic information. SeLRoS is designed with a modular architecture, comprising geometric room segmentation, object mapping, and semantic integration components. geometric room segmentation forms the foundational layer of SeLRoS, where the 2D map extracted from the original environment undergoes segmentation using the VRF algorithm. object mapping then creates object information based on visual data obtained from the center coordinates of each segmented room. semantic integration represents the culmination of the entire process, where the segmented spaces are merged with object information and spacial context information. This phase leverages LLMs to interpret the contextual relationships between objects and spaces, generating the semantic information that used to enhance segmentation map. To evaluate the effectiveness of SeLRoS, we conducted extensive experiments across 30 diverse environments generated by AI2-THOR, analyzing the results through both qualitative and quantitative analysis. These experiments showed the capabilities of SeLRoS in improving room segmentation accuracy by leveraging semantic insights.

Despite its promising outcomes, our study has several limitations. Initially, while SeLRoS can effectively integrate segmented rooms within the segmentation map, it cannot divide spaces for more detailed refinement. This limitation impacts the system's ability to fine-tune room boundaries for improved accuracy. Additionally, SeLRoS determines the center of each segmented room using a scalar mean to collect visual data. Whether these central points best capture the essence of each room requires further discussion and investigation. Finally, using ProcTHOR from AI2-THOR to create indoor environments sometimes presents challenges because the automatically generated structures can be placed in unusual or mismatched contexts in a few cases. These situations can make it hard for SeLRoS, which relies on understanding the context, to add correct semantic information to the segmentation results.

Therefore, future research will aim to more systematically integrate segmentation maps with a broader range of contextual information. This goal will enhance the accuracy of both semantic information and segmentation maps, resulting in a more precise representation of indoor environments.

\bibliographystyle{IEEEtran}
\bibliography{references}
\end{document}